\documentclass[a4paper,10pt]{article}

\usepackage{ucs}
\usepackage[utf8x]{inputenc}

\newcommand{\pname}{{\sc DAG}itty\xspace}
\newcommand{\action}[1]{\emph{``#1''}}

\title{Drawing and Analyzing Causal DAGs with \pname}
\author{Johannes Textor}
\date{August 19th, 2015}

\usepackage{atbegshi}
\usepackage{tikz}
\usepackage{a4wide}
\usepackage{xspace}
\usepackage{hyperref}
\usepackage{pxfonts}

\usepackage[nottoc,numbib]{tocbibind}

\begin{document}

\maketitle

\begin{abstract}
\pname is a software for drawing and analyzing causal 
diagrams, also known as directed acyclic graphs (DAGs).
Functions include identification of minimal sufficient
adjustment sets for estimating causal effects, 
diagnosis of insufficient or invalid adjustment via the identification
of biasing paths, identification of instrumental variables, 
and derivation of testable implications.

\pname is provided in the hope that it is useful for researchers
and students in Epidemiology, Sociology, Psychology, and
other empirical disciplines. The software should run in any modern
web browser that supports JavaScript, HTML, and SVG.

This is the user manual for \pname version 2.3. The manual is updated with every 
release of a new stable version. \pname is available at 
\href{http://dagitty.net}{\tt dagitty.net}.
\end{abstract}

\tableofcontents

\section{Introduction}

\pname is a web-based software for analyzing causal diagrams. It contains some of the 
fastest algorithms available for this purpose. 

This manual describes how causal diagrams can be created
(Section~\ref{sec:diagramcreation}) 
and manipulated (Section~\ref{sec:diagramediting}) using \pname. In 
Section~\ref{sec:diagramanalysis}, \pname's capabilities to analyze causal diagrams are
described. A brief introduction to causal diagrams is given
in Section~\ref{sec:dagintro}.

\subsection{Citing \pname}

Developing and maintaining \pname requires a substantial
amount of work; thus, if you publish research results obtained
with the help of \pname, please 
consider giving us credit by citing our work. 
Depending on the context, you could cite the 
letter in {\em Epidemiology} where \pname has first been announced \cite{Textor2011},
or the research papers describing the specific algorithms used
to identify biasing paths \cite{TextorL2011}, 
adjustment sets \cite{textor14_uai},
and instrumental variables \cite{zander15_instrument}.

\subsection{Running \pname online}

There are two ways to run \pname: either from the internet or from your
own computer. To run \pname online, visit the URL
\href{http://www.dagitty.net}{\tt dagitty.net}.
\pname should run in every modern browser. Specifically, I expect it
to work well on recent versions of Firefox, Chrome, Opera, and Safari
as well as on Internet Explorer (IE) version 9.0 or later, which all
support scalable vector graphics (SVG). IE
versions prior to 9.0 do not support SVG. These should be able to perform
all diagnosis functions but cannot display the graphics as well as modern
browsers can\footnote{While this would be redeemable, I'd much rather invest my
time in improving \pname for modern browsers than fixing it for old IE versions.
If you absolutely need to run \pname on older IEs and encounter severe problems,
please contact me.}. If you encounter any problems, please send
me an e-mail so I can fix them (my contact information is at the end of this manual). 
Keep in mind that \pname is often used by hundreds of people per day
from all over the world -- these people all benefit if the problem you found is 
fixed so please do consider investing
the time to notify me if you encounter any bugs.

\subsection{Installing \pname on your own computer} 

\pname can be ``installed'' on your computer for use without an internet
connection. To do this, download the file 
\href{http://www.dagitty.net/dagitty.zip}{\tt dagitty.net/dagitty.zip}
which is a ZIP archive containing \pname's source. Unpack this ZIP file anywhere
in your file system. To run \pname, just open the file \verb|dags.html| in the
unpacked folder. 

Some features of \pname will not work in the offline version, because they
are actually implemented on the web server. Currently, these features are:
\begin{itemize}
\item Exporting model drawings as PDF, JPEG or PNG files.
\item Publishing models on-line.
\end{itemize}

\subsection{Migrating from earlier versions of \pname}

The following two issues are important for users of older \pname versions. New users
can skip this section.

\begin{itemize}
 \item It is now possible to have more than one exposure and/or outcome variable. 
This means that the old model code convention where the variable in the first line is the 
exposure and the variable in the second line is the outcome no longer works. Hence,
if you open a model created with an earlier version in \pname 2.0, 
exposure and outcome will appear like normal variables. To fix this, 
simply set exposure and outcome again
using the \action{e} and \action{o} keys  and save the new model code.

 \item Spaces in variables are now finally reliably supported. The way this works
is that any variable name containing spaces or other special symbols is stored
using ``URL encoding'' -- e.g. ``patient sex'' will turn into ``patient\%20sex''
(of course \pname will do this automatically for you). 
This may look strange but ensures that \pname models can be safely e-mailed, posted
on websites, stored in word documents and so forth without having to worry about line 
breaks messing up variable names. 
If you have an older \pname model containing spaces in variable names,
\pname 2.0 or higher should open this model correctly and perform the conversion itself. If
it does not, consider sending me your model so I can investigate.
\end{itemize}

\section{A brief introduction to causal diagrams}

\label{sec:dagintro}

In this section, we will briefly review what causal diagrams are and how they can be 
applied in empirical sciences. For a more detailed account, we recommend 
the book \emph{Causality} by Judea Pearl \cite{Pearl2009}, or the 
chapter \emph{Causal Diagrams} in the Epidemiology 
textbook of Rothman, Greenland, and Lash \cite{RothmanGL2008}. 
Also take a look at the web page 
\href{http://www.dagitty.net/learn/}{\tt dagitty.net/learn/},
where I am collecting several tutorials (some of them interactive) on specific DAG-related 
topics.
 
In Epidemiology, causal diagrams are also frequently
called \emph{DAGs}.\footnote{The term ``DAG'' is
somewhat confusing to computer scientists and mathematicians, 
for whom a DAG is simply an abstract mathematical structure without specific semantics 
attached to it.} In a nutshell, a DAG is a graphic model that depicts a set of hypotheses
about the causal process that generates a set of variables of interest.
An arrow $X \to Y$ is drawn if there is a direct
causal effect of $X$ on $Y$. Intuitively, this means that
the natural process determining $Y$ is directly influenced by
the status of $X$, and that altering $X$ via external intervention 
would also alter $Y$. However, an arrow $X \to Y$ only represents that part
of the causal effect which is \emph{not} mediated by any of the other variables
in the diagram. If one is certain that $X$ does not
have a direct causal influence on $Y$, then the arrow is omitted.
This has two important implications: (1) arrows should follow time order,
 or else the diagram contradicts the basic principle that
causes must precede their effects; (2) the omission of an arrow is a stronger
claim than the inclusion of an arrow -- the presence of an arrow depicts
merely the ``causal null hypothesis'' that $X$ \emph{might} have an effect
on $Y$.

Mathematically, the semantics of an arrow $X \to Y$ can be defined as
follows. Given a DAG $G$ and a variable $Y$ in $G$, let $X_1,\ldots,X_n$
be all variables in $G$ that have direct arrows $X_i \to Y$ (also called the
\emph{parents} of $Y$). Then $G$ 
claims that the causal process determining the value of $Y$ can be
modelled as a mathematical function $Y := f(X_1,\ldots,X_n,\epsilon_Y)$,
where $\epsilon_Y$ (the ``causal residual'') is a random variable that
is jointly independent of all $X_i$.  

For example, the sentence ``smoking causes lung cancer'' could be translated
into the following simple causal diagram: 

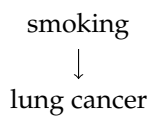
\begin{center}
\begin{tikzpicture}
\node (sm) {smoking};
\node (c) [below of=sm] {lung cancer};

\draw [->] (sm) -- (c);
\end{tikzpicture}
\end{center}

We would interpret this diagram as
follows: (1) The variable ``smoking'' refers to a person's smoking habit
prior to a later assessment of cancer in that same person;
(2) the natural process by which a person develops
cancer might be influenced by the smoking habits of that person; (3) there
exist no other variables that have a direct influence on both smoking 
habits and cancer. A slightly
more complex version of this diagram might look as follows:

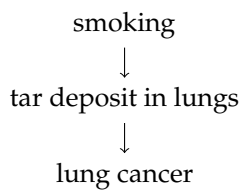
\begin{center}
\begin{tikzpicture}
\node (sm) {smoking};
\node [below of=sm] (t) {tar deposit in lungs};
\node (c) [below of=t] {lung cancer};

\draw [->] (sm) -- (t);
\draw [->] (t) -- (c);
\end{tikzpicture}
\end{center}

This diagram is about a person's smoking habits at a time $t_1$,
the tar deposit in her lungs at a later time $t_2$, and finally the
development of lung cancer at an even later time $t_3$. We claim that 
(1) the natural process which determines the amount of tar in the
lungs is affected by smoking; 
(2) the natural process by which lung cancer develops is affected by
the amount of tar in the lung; (3) the natural 
process by which lung cancer develops is not affected by the person's
smoking other than indirectly via the tar deposit; 
and finally (4) no variables having
relevant direct influence on more than one variable
of the diagram were omitted.

In an epidemiological context, we are often interested in the putative 
effect of a set of variables, called \emph{exposures}, on another set of variables
called \emph{outcomes}. A key question in Epidemiology (and many other empirical sciences)
is: how can we infer the causal effect of an exposure on an outcome of interest from
an observational study? Typically, a simple regression will not suffice due to the 
presence of \emph{confounding factors}. 
If the assumptions encoded in a given the diagram hold,
then we can infer from this diagram sets of variables 
for which to adjust in an observational study to 
minimize such confounding bias. For example,
consider the following diagram: 

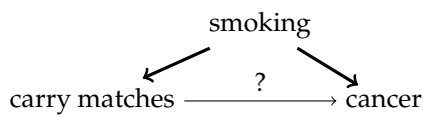
\begin{center}
\begin{tikzpicture}
\node (sm) at (0,0)  {smoking};
\node [anchor=east] (m) at (-1,-1) {carry matches};
\node [anchor=west] (c) at (1,-1) {cancer};
\draw [->,very thick] (sm) -- (c);
\draw [->,very thick] (sm) -- (m);
\draw [->] (m) -- node[midway,above] {?} (c);
\end{tikzpicture}
\end{center}

If we were to perform an association 
study on the relationship between carrying matches
in one's pocket and developing lung cancer, we would probably find a 
correlation between these two variables. However, as the above diagram
indicates, this correlation would not imply that carrying matches in 
your pocket causes lung cancer: Smokers are more likely to carry matches
in their pockets, and also more likely to develop lung cancer. This is an
example of a \emph{confounded} association between two variables,
which is mediated via the \emph{biasing path} (bold). 
In this example, let us assume with a leap of faith that the simplistic diagram
above is accurate. Under this assumption, would we adjust for smoking, 
e.g. by averaging separate effect estimates for 
smokers and non-smokers, we would  
no longer find a correlation between carrying matches and lung cancer.
In other words, adjustment for smoking would \emph{close the biasing path}.
Adjustment sets will be explained in more detail in 
Section~\ref{sec:adjustment}.

The purpose of \pname is to aid study design through the identification of 
suitable, small sufficient adjustment sets in complex causal diagrams and,
more generally, through the identification of causal and biasing paths as
well as testable implications in a given diagram. 

\section{Loading, saving  and sharing diagrams}

\label{sec:diagramcreation}

This section covers the three basic steps of working with \pname: 
(1) loading a diagram; (2) manipulating the graphical layout of the diagram; and
(3) saving the diagram. 
First of all, any causal diagram consists of vertices (variables) and arrows 
(direct causal effects).
You can either create the diagram directly using \pname's graphical user
interface (explained in the next section),
or prepare a textual diagram description in a word processor 
and then import this
description into \pname. In addition, \pname
contains some pre-defined examples that you can use to become familiar
with the program and with DAGs in general. 
To do so, just select one of the pre-defined examples from
the \action{Examples} menu.

\subsection{\pname's textual syntax for causal diagrams} 

\pname's textual syntax for causal diagrams is based on the one used
by the DAG program by Sven Kn\"{u}ppel \cite{KnueppelS2010}. A diagram description 
(\emph{model code}, somewhat clumsily 
called \emph{model text data} in older \pname versions)  
consists of two parts: 

\begin{enumerate}
  \item A list of the variables in the diagram
  \item A list of arrows between the variables
\end{enumerate}

The list of variables consists of one variable per line. 
After each variable name follows 
a character that indicates the status of the variable, which can
be one of ``1'' (normal variable), ``A'' (adjusted for), ``U'' (latent/unobserved),
``E'' (exposure), or ``O'' (outcome). If you prepare your diagram description
in a word processor rather than constructing the diagram in \pname itself,
you may encounter problems when you use spaces or other special symbols
in variable names (e.g. instead of ``patient sex'' you should write 
``patient\_sex''). This restriction does not apply when you construct the
diagram using \pname's graphical user interface.

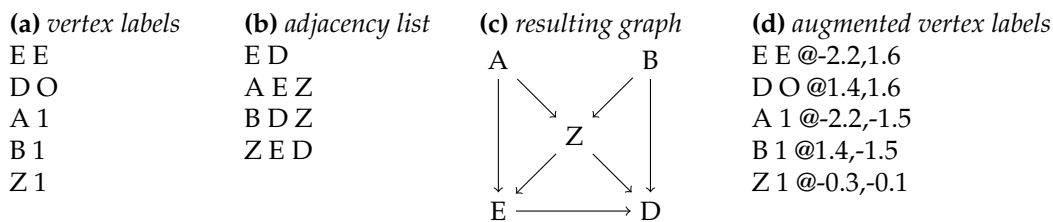
\begin{figure}
\centering 
\begin{minipage}[t]{3cm}
{\bf (a)} \emph{vertex labels}\\
E E \\
D O \\
A 1 \\
B 1 \\
Z 1
\end{minipage}
\begin{minipage}[t]{3cm}
{\bf (b)} \emph{adjacency list}\\
E D \\
A E Z \\
B D Z \\
Z E D
\end{minipage}
\begin{minipage}[t]{3cm}
{\bf (c)} \emph{resulting graph}\\
\begin{tikzpicture}
\node (e) at (-1,0) {E};
\node (d) at (1,0) {D};
\node (z) at (0,1) {Z};
\node (a) at (-1,2) {A};
\node (b) at (1,2) {B};
\draw [->] (a) -- (z); \draw [->] (b) -- (z); 
\draw [->] (a) -- (e); \draw [->] (b) -- (d); 
\draw [->] (z) -- (e); \draw [->] (z) -- (d);
\draw [->] (e) -- (d);
\end{tikzpicture}
\end{minipage}
\hskip .5cm
\begin{minipage}[t]{4cm}
{\bf (d)} \emph{augmented vertex labels}\\
E E @-2.2,1.6 \\
D O @1.4,1.6 \\
A 1 @-2.2,-1.5 \\
B 1 @1.4,-1.5 \\ 
Z 1 @-0.3,-0.1
\end{minipage}
\caption{Example for a textual model definition with \pname (a,b: model code; 
c: resulting diagram). When the diagram is edited within
\pname, the vertex labels and adjustment status are augmented with 
additional information that \pname uses to layout 
the vertices on the canvas (d): the
layout coordinates of each variable are indicated behind the @ sign. 
} 
\label{fig:syntaxexample}
\end{figure}

The list of arrows consists of several lines each starting 
with a start variable name, followed by one or more other target variables 
that the start variable is connected to. Figure~\ref{fig:syntaxexample} 
contains a worked example of a textual model description. When you modify
a diagram within \pname, the vertex labels will be augmented by additional information
to help \pname remember the layout of the vertices and for other purposes (see
rightmost column in Figure~\ref{fig:syntaxexample}).

\subsection{Loading a textually defined diagram into \pname}

To load a textually defined diagram into \pname, simply copy\&paste the 
variable list, followed by a blank line,
followed by the list of arrows into the \action{Model code} text
box. Then click on \action{Update DAG}.
\pname will now generate a preliminary graphical layout for your diagram on 
the canvas, which may not yet look the way you intended, but can be freely 
modified.

\subsection{Modifying the graphical layout of a diagram}

To layout the vertices and arrows of your diagram more clearly than \pname
did, simply drag the vertices with your mouse on the canvas. You may notice
that \pname modifies the information in the \action{Model code} field
on the fly, and augments it with additional position information for each
vertex. In general, all changes you make to your diagram within \pname 
are immediately reflected in the model code. 

\subsection{Saving the diagram} 

To save your diagram locally, just copy\&paste the contents of the \action{Model code}
field to a text file, 
and save that file locally to your computer\footnote{This is most easily done
by clicking in the text field, pressing \action{CTRL + A} to select the entire content
of the text field, then pressing \action{CTRL + C} to copy the content. You
can then paste the content in another program using \action{CTRL + V}.}.
When you wish to continue working on the
diagram, copy the model code back into \pname as explained above. 

\subsection{Exporting the diagram}

\pname can export the diagram as a PDF or SVG vector graphic (publication quality)
or a JPEG or PNG bitmap graphic (e.g. for inclusion in Powerpoint). Select the
corresponding function from the \action{Model} menu. If you want to edit the graphical
layout of the diagram or annotate it, it is recommended to export the diagram as an
SVG file and open that in a vector graphics program such as Inkscape.

\subsection{Publishing diagrams online}

Part of the appeal of using DAGs is that the assumptions underlying
one's research are made explicit, and the conclusions drawn from 
the data can be later re-checked if some of the assumptions are found
to not hold. Of course, this requires to make the DAG available together
with the data and interpretation. I have however seen many articles
where people report having used DAGs but do not actually show them.
If researchers, reviewers or editors deem it inappropriate to include the 
DAG (or its model code) in the manuscript itself, here's another option:
Store the DAG on the \pname website and get a short URL under which
this DAG will be accessible. Then include this URL in the manuscript,
or its supporting information. For example, one of the \pname examples
is stored at the URL \href{http://dagitty.net/mvcFQ}{\tt dagitty.net/mvcFQ}.

Here's how it works: Draw your DAG to full satisfaction, then choose 
\action{Publish on dagitty.net} from the \action{Model} menu. You have
two options how to publish your DAG: anonymously, or linking it to an e-mail
address. If you store the DAG anonymously, you will later on not be able
to edit it or delete it from the server.

After choosing \action{Publish on dagitty.net} from the \action{Model} menu,
a small form will appear where you can enter some metadata on the DAG,
and provide your e-mail address if you so wish. Upon clicking \action{Publish},
the DAG will be sent to the dagitty.net server, and you will receive a URL under
which the DAG is now available. If you provided your e-mail address, you
will also receive a message requesting you to confirm your ownership of the DAG.
This is simply done by clicking on a confirmation link. Only then will the DAG
be linked to your e-mail address, and you will receive a password to use when
deleting or modifying the published DAG.

If you did link your DAG to your e-mail address, you can delete it by choosing
\action{Delete on dagitty.net} from the \action{Model} menu, which will prompt you
to enter the DAG's URL and the password. If the URL and password match,
the DAG will be deleted.
Similarly, you can update a stored DAG using the \action{Load from dagitty.net} function
from the  \action{Model} menu, modifying it, and saving it again. 
You can view published DAGs (if you know their URL)
by just putting the URL into your address bar of course, but you can also do
so using the \action{Load from dagitty.net} function. 

Please note that all DAGs stored on dagitty.net are meant to be public information.
Do not store any data that you consider private or in any way secret.
Once stored on dagitty.net, every person in the world who knows your DAG's URL
can view it (but not your e-mail address if you provided one). Also note that there is
no guarantee that dagitty.net will keep running forever. Storing your DAGs is done at
your own risk. Still, you may find this feature useful, for instance to e-mail your DAGs
to colleagues or to include links to DAGs in papers under review. For archival
purposes, it may be more appropriate to include the DAG or the model code in
the paper itself or its supporting information.

\section{Editing diagrams using the graphical user interface} 

\label{sec:diagramediting}

You are free to make changes directly to the textual description of your
diagram, which will be reflected on the canvas next time you click on \action{Update DAG}.
However, you can also create, modify, and delete vertices and arrows graphically
using the mouse.

\subsection{Creating a new diagram}

To create a new diagram, select \action{New Model} from the \action{Model} menu.
You will be asked
for the names of the exposure and the outcome variable, and an initial model containing
just those variables and an arrow between them will be drawn. Then you can
add variables and arrows to the model as explained below. 

\subsection{Adding new variables}

To add a new variable to the model, double-click on a free space in the canvas
(i.e., not on an existing variable) or press the \action{n} key. A dialog will
pop up asking you for the name of the new variable. Enter the name into the dialog
and press the enter key or click \action{OK}. If you click \action{Cancel}, no new 
variable will be created.

\subsection{Renaming variables}

To rename an existing variable, move the
mouse pointer over that variable and hit the \action{r} key. A dialog
will pop up allowing you to change the variable name.

\subsection{Setting the status of a variable}

Variables can have one of the following 
statuses:

\begin{itemize} 
\item Exposure
\item Outcome
\item Unobserved (latent)
\item Adjusted
\item Other
\end{itemize}

To turn a variable into an exposure, move the
mouse pointer over that variable and hit the \action{e} key; 
for an outcome, hit the \action{o} key instead.
To toggle whether a variable is observed or unobserved,
hit the \action{u} key; to toggle whether it is 
adjusted, hit the \action{a} key.
Changing the status of variables may change the 
colors of the diagram vertices to reflect the new structure 
and information flow in the diagram (see below). 

At present, these statuses are mutually exclusive --  e.g.,
a variable cannot be both unobserved and adjusted 
or both exposure and unobserved. This could change
in future versions of \pname.

\subsection{Adding new arrows}

To add a new arrow, double-click first on the source vertex 
(which will become highlighted) and then on the target vertex.
The arrow will be inserted. If an arrow existed before
in the opposite direction, that arrow will be deleted, because otherwise
there would now be a cycle in the model. 

Instead of double-clicking on a vertex, you can also move the mouse pointer
over the vertex and press the key \action{c}. Arrows are by default
drawn using a straight line,
but you can change that moving the mouse pointer to the line,
pressing and holding down the left mouse button, and ``bending''
the line by dragging as appropriate.

\subsection{Deleting variables}

To delete a variable, move the mouse pointer over that variable 
and hit the \action{del} key on your keyboard, or alternatively
the \action{d} key (the latter comes in handy if you're on a Mac, which 
has no real delete key).
All arrows to that variable will be deleted along 
with the variable. In contrast to \pname versions prior to 2.0,
all variables can now be deleted including exposure and outcome.

\subsection{Deleting arrows}

An arrow is deleted just like it has been inserted, 
i.e., by double-clicking first on the start 
variable and then on the target variable. An arrow 
is also deleted automatically if a new one is 
inserted in the opposite direction (see above). 

\subsection{Choosing the style of display}

At present, you can choose between two DAG diagram styles: ``classic'', where nodes and
their labels are separate from each other, and SEM-like, where labels are inside nodes. 
Both have their advantages and disadvantages. By the way, ``SEM'' refers to structural
equation modeling.

\section{Analyzing diagrams}

\label{sec:diagramanalysis}

\subsection{Paths}

Causal diagrams contain two different kinds of paths between exposure and outcome
variables.

\begin{itemize}
\item \emph{Causal paths} start at the exposure, contain
only arrows pointing away from the exposure, and end at 
the outcome. That is, they have the form
$e \rightarrow x_1 \rightarrow \ldots \rightarrow x_k \rightarrow o$.
\item \emph{Biasing paths} are all other paths from exposure to outcome. For 
example, such paths can have the form 
$e \leftarrow x_1 \rightarrow \ldots \rightarrow x_k \rightarrow o$.
\end{itemize}

With respect to a set $\mathbf{Z}$ of conditioning variables (that 
can also be empty if we are not conditioning on anything), 
paths can be either \emph{open} or \emph{closed}
(also called d-separated \cite{Pearl2009}). A path 
is \emph{closed} by $\mathbf{Z}$ if one or both of the following holds: 

\begin{itemize}
 \item The path $p$ contains a chain $x \rightarrow m \rightarrow y$ 
 	or a fork $x \leftarrow m \rightarrow y$ 
    such that $m$ is in $\mathbf{Z}$.
 \item The path $p$ contains a collider $x \rightarrow c \leftarrow y$ 
 	such that $c$ is not in $\mathbf{Z}$ and 
    furthermore, $\mathbf{Z}$ does not contain
    any successor of $c$ in the graph. 
\end{itemize}

Otherwise, the path is \emph{open}.
The above criteria imply that paths consisting of only one
arrow are always open, no matter the content of $\mathbf{Z}$. 
Also it is possible that a path is closed with respect
to the empty set $\mathbf{Z}=\{\}$.

\subsection{Coloring}

It is not easy to verify by hand which paths are open and
which paths are closed, especially in larger diagrams. 
\pname highlights all arrows lying on open 
biasing paths in red and all arrows lying on open
causal paths in green. This highlighting is optional 
and is controlled via 
the \action{highlight causal paths} and \action{highlight biasing paths}
checkboxes.

\subsection{Effect analysis}

As mentioned above, arrows in DAGs represent \emph{direct effects}. That is, in a DAG
with three variables $X$, $M$, and $Y$, an arrow $X \to Y$ means that there is a causal
effect of $X$ on $Y$ that is \emph{not} mediated through the variable $M$. 
Often when building DAGs, people tend to forget this aspect and think only about whether
any kind of causal effect exists, without paying attention to how it is mediated. This 
may result in DAGs with too many arrows. 

To aid users with this, George Ellison (Leeds University) suggested to implement a 
function that identifies arrows for which also a corresponding indirect pathway exists. 
After drawing an initial DAG, one might reconsider these arrows and judge whether they
are really necessary given the indirect pathways already present in the diagram.

For example, suppose after thinking about the pairwise causal relationships between
our variables $X$, $M$, $Y$ we came up with this DAG:

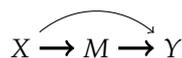
\begin{tikzpicture}
\node (x) at (0,0) {$X$};
\node (m) at (1,0) {$M$};
\node (y) at (2,0) {$Y$};
\draw [very thick,->] (x) -- (m);
\draw [very thick,->] (m) -- (y);
\draw [bend left=40,->] (x) edge (y);
\end{tikzpicture}

For the arrows drawn in bold, there is no corresponding indirect path -- removing
one of these arrows from the diagram means that there will no longer be any causal 
effect between the corresponding variables. These arrows are called \emph{atomic direct
effects} in \pname, and they can be highlighted -- like in the above DAG -- by ticking
the checkbox with that name. 
On the other hand, for the thin arrow 
$X \to Y$, there is also the indirect pathway $X \to M \to Y$. One may therefore reconsider
whether the arrow $X \to Y$ is truly necessary -- 
perhaps the causal effect from $X$ to $Y$ is entirely mediated through $M$.

\subsection{View mode}

There are several ways to transform a given DAG such that it becomes better suited for 
a particular purpose. We call such a transformed DAG a \emph{derived graph}. Currently
\pname can display two kinds of derived graphs: correlation graphs, and moral graphs.
These derived graphs can be shown by clicking on the respective radio button in the
\action{View mode} field on the left-hand side of the screen. 

\subsubsection{The correlation graph}

The correlation graph is not a DAG, but a simple graph with lines instead of arrows. 
It connects each pair of variables that, according to the diagram, could be statistically
dependent. In other words, variables not connected by a line in the correlation graph
must be statistically independent. These pairwise independencies are also listed in the
\action{Testable implications} field on the right-hand side of the screen, and so the
correlation graph could be seen as encoding a subset of those implications. 

Although this is not implemented in \pname yet, it is also possible to take a given 
correlation graph (which can be obtained e.g. by thresholding a covariance matrix) and 
list all the DAGs that are ``compatible'' with it in the sense that they entail exactly
the given correlation graph \cite{textor15_uai}.

\subsubsection{The moral graph}

To identify minimal sufficient adjustment sets, \pname uses the so-called ``moral graph'',
which results from a transformation of the model to an undirected graph.
This procedure is also highly recommended if you wish to verify the calculation by hand.
See the nice explanation by Shrier and Platt \cite{ShrierP2008} for details on this
procedure.  

In \pname, you can switch between display of the model and its moral graph 
choosing ``moral graph'' in the``view mode'' section on the left-hand
side of the page.

\subsection{Causal effect identification}

Some of the most important features of \pname are concerned with the question: how
can causal effects be estimated from observational data? Currently, two types of causal
effect identification are supported: adjustment sets, and instrumental variables.

\subsubsection{Adjustment sets}

\label{sec:adjustment}

Finding sufficient adjustment sets is one main purpose 
of \pname. In a nutshell, a sufficient adjustment
set $\mathbf{Z}$ is a set of covariates such that adjustment,
stratification, or selection (e.g. by restriction or 
matching) will minimize bias when estimating the causal 
effect of the exposure on the outcome (assuming that
the causal assumptions encoded in the diagram hold). 
You can read more about controlling bias and confounding in Pearl's textbook, 
chapter 3.3 and epilogue \cite{Pearl2009}. 
Moreover, Shrier and Platt \cite{ShrierP2008} 
give a nice step-by-step tutorial on how to test if a set of covariates
is a sufficient adjustment set.

To identify adjustment sets, the diagram must contain at least
one exposure and at least one outcome.

\paragraph{Total and direct effects.}
One can understand adjustment sets graphically by viewing an adjustment set
as a set $\mathbf{Z}$ that closes all all biasing 
paths while keeping desired causal paths open
(see previous section).
\pname considers two kinds of adjustment sets: 

\begin{itemize}
 \item Adjustment sets for the \emph{total effect} are sets that 
close all biasing paths and leave all causal paths
open. 
 In the literature, if the effect is not mentioned
(e.g. \cite{ShrierP2008,KnueppelS2010}), then usually
this kind of adjustment set is meant.
 \item   Adjustment sets for the \emph{direct effect} are sets that 
close all biasing paths and all causal paths, and leave only 
the direct arrow from exposure $X$ to outcome $Y$ (i.e., the path
$X \rightarrow Y$, if it exists) open.
\end{itemize}

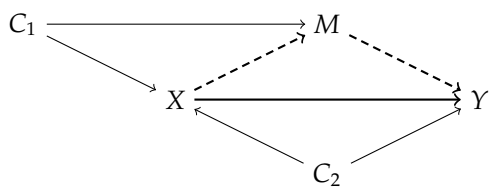
\begin{figure}
\begin{center}
\begin{tikzpicture} [scale=2]
\tikzstyle{vertex}=[rectangle] 
\node at(-1,.5)(C1) [vertex]{$C_{1}$};
\node at(1,-.5)(C2)[vertex]{$C_{2}$};
\node at(2,0) (Y)  [vertex]{$Y$}
	edge [<-] (C2);
\node at(1,.5) (M)  [vertex]{$M$}
	edge [->,thick,densely dashed] (Y)
	edge [<-] (C1);	
\node at(0,0) (X)  [vertex]{$X$}
	edge [->,thick,densely dashed] (M)
	edge [<-] (C1)
	edge [<-] (C2);	
\draw (X) edge [->,thick] (Y);
\end{tikzpicture}
\end{center}

\caption{A causal diagram where the total and direct 
effects of exposure $X$ on outcome $Y$ are not equal. The total effect is the effect
mediated only via the thick (both dashed and solid) arrows,
while the direct effect is the effect mediated only 
via the thick arrow.}
\label{fig:effects}

\end{figure}

In a diagram where the only causal path between exposure 
and outcome is the path $X \rightarrow Y$, the total effect
and the direct effect are equal. This is true e.g. for the 
diagram in Figure~\ref{fig:syntaxexample}. An example diagram where the 
direct and total effects are not equal is shown in 
Figure~\ref{fig:effects}.

As proved by Lauritzen et al. \cite{Lauritzen1990} 
(see also Tian et al. \cite{TianPP1998}), 
it suffices to restrict our attention to the part of the model that consists of exposure, 
outcome, and their ancestors for identifying sufficient adjustment sets. This is indicated
by \pname by coloring irrelevant nodes in gray. The relevant variables are colored according 
to which node they are ancestors of (exposure, outcome, or both) -- see the legend on the 
left-hand side of the screen. The highlighting may be turned on and off by toggling the 
\action{highlight ancestors} checkbox.

\paragraph{Minimal sufficient adjustment sets.} 

A \emph{minimal} sufficient adjustment set is a sufficient 
adjustment set of which no proper subset is itself sufficient. For example, 
consider again the causal diagram in Figure~\ref{fig:syntaxexample}. 
The following three sets are sufficient adjustment sets
for the total and direct effects, which are equal in this case: 

$$ \{A,B,Z\} $$

$$ \{A,Z\} $$

$$ \{B,Z\} $$ 

Each of these sets is sufficient because it closes all biasing paths and leaves 
the causal path open. The sets $\{A,Z\}$ and $\{B,Z\}$ are minimal sufficient adjustment 
sets while the set $\{A,B,Z\}$ is sufficient, but not minimal. 
In contrast, the set $\{Z\}$ is \emph{not} sufficient, since this would 
open  
the path $E \leftarrow A \rightarrow Z \leftarrow B \leftarrow D$: 
Because both $E$ and 
$D$ depend on $Z$, adjusting for $Z$ will induce 
additional correlation between $E$ and $D$. 

\paragraph{Finding minimal sufficient adjustment sets.}

To find minimal sufficient adjustment sets, select the 
option \action{Adjustment (total effect)}
or \action{Adjustment (direct effect)} in the 
\action{Causal effect identification} field. 
\pname
will then calculate all minimal sufficient adjustment sets and display 
them in that field. Any changes made to the diagram will be instantly
reflected in the list of adjustment sets.

\paragraph{Forcing adjustment for specific covariates.}

You can also tell \pname that you wish a specific covariate
to be included into every adjustment set. To do this, move the
mouse over the vertex of that covariate and press the \emph{a} 
key. \pname will then update the list of minimal sufficient
adjustment sets accordingly -- every set displayed
is now minimal in the sense that removing any variable 
\emph{except those you specified} will render
that set insufficient. However, when you adjust for an intermediate
or another descendant of the exposure, \pname will tell you that it is no longer
possible to find a valid adjustment set.

\paragraph{Avoiding adjustment for unobserved covariates.}

You can tell \pname that a certain variable is unobserved 
(e.g. not measured at present, or not measurable because it is a 
latent variable) by moving the mouse 
over that covariate and pressing the \emph{u} key.
\pname will only calculate adjustment sets that do 
not contain unobserved variables. However, if too many or
some important variables are unobserved, then it may
be impossible to close all biasing paths. 

\subsubsection{Instrumental variables}

Sometimes it is not possible to estimate a causal effect by simple covariate adjustment.
For example, this is the case whenever there is an unobserved confounder that directly
effects the exposure and outcome variables. However, this does not necessarily 
mean that it is impossible to estimate the causal effect at all. 
\emph{Instrumental variable regression} is a technique that is 
often used in situations with unobserved confounders.
Note that this technique depends on linearity assumptions. For further information
on instrumental variables, please refer to the literature \cite{AngristIR96,imbens14}. 
\pname can find instrumental variables in DAGs, as explained below. 

The validity of an instrumental variable $I$ depends on two causal conditions: 
exogeneity, and exclusion restriction. These two conditions can be expressed 
in the language of DAGs and paths as follows: (1) there must be an open path between 
$I$ and the exposure $X$; and (2) all paths between $I$ and the outcome $Y$ must be
 closed by $\{X\}$.
A variable that fulfills these two conditions is called an \emph{instrumental variable}
or simply an \emph{instrument}.

Instrumental variables can also be generalized such that the two conditions are required
to hold \emph{conditional on a set of covariates} $\mathbf{Z}$ \cite{BritoPearlUAI02}. 
The two conditions then read as follows: 
(1) there must be a path between 
$I$ and $X$ that is opened by $\mathbf{Z}$; and (2) 
all paths between $I$ and $Y$ must be closed by $\mathbf{Z} \cup \{X\}$.
A variable that fulfills these two conditions is called a
\emph{conditional instrument}.

\pname will find both ``classic'' and conditional instruments when the option 
\action{Instrumental Variable} is selected under the \action{Causal effect identification} 
field. Note that \pname will not always list \emph{all} possible instruments;
instead, it will restrict itself to a certain well-defined subset that we call 
``ancestral instruments''. However, whenever any instrument or conditional instrument
exists at all, then \pname is guaranteed to find one. Note also that if there are several
instruments available, then it is best to choose the one that is most strongly
correlated with $X$ (conditional on $\mathbf{Z}$ in the case of a conditional instrument).

For details regarding ancestral instruments and how \pname computes them, please
refer to the research paper where we describe these methods \cite{zander15_instrument}.

\subsection{Testable implications}

Any implications that are obtained from a causal diagram,
such as possible adjustment sets or instrumental variables,
are of course dependent on the 
assumptions encoded in the diagram.
To some extent, these assumptions can be tested via the 
(conditional) independences implied by the diagram: If two 
variables $X$ and $Y$ are $d$-separated by a set $\mathbf{Z}$, then
$X$ and $Y$ should be conditionally independent
given $\mathbf{Z}$. The converse is not true: Two variables
$X$ and $Y$ can be independent given a set $\mathbf{Z}$ even though they
are not $d$-separated in the diagram. Furthermore, two 
variables can also be $d$-separated by the empty set $\mathbf{Z}=\emptyset$.
In that case, the diagram implies that $X$ and $Y$ are
\emph{unconditionally} independent.

\pname displays all minimal testable implications 
in the \action{Testable implications} text field. Only such
implications will be displayed that are in fact testable,
i.e., that do not involve any unobserved variables. Note
that the set of testable implications displayed by \pname
does not constitute a ``basis set'' \cite{Pearl2009}. 
Future versions will allow choosing between different basis sets. 

In general, the less arrows a diagram contains, the more testable
predictions it implies. For this reason, ``simpler'' models with
fewer arrows are in general easier to falsify (Occam's razor).

\section{Acknowledgements}

I would like to thank my collaborators Maciej Li\'skiewicz
and Benito van der Zander (both at the Institute for Theoretical Computer Science,
University of L\"ubeck, Germany) for our collaborations on developing efficient algorithms
to analyze causal diagrams.  

I also thank Michael Elberfeld, Juliane Hardt, 
Sven Kn\"{u}ppel, Keith Marcus, Judea Pearl, Sabine Schipf, 
and Felix Thoemmes (in alphabetical order) for enlightening discussions
(either in person, per e-mail, or on the SEMnet discussion list)
about DAGs that made this program possible. Furthermore, I thank
Robert Balshaw, George Ellison, 
Marlene Egger, Angelo Franchini,  Ulrike F\"{o}rster, Mark Gilthorpe, 
Dirk van Kampen,
Jeff Martin, Jillian Martin, Karl Micha\"{e}lsson,
David Tritchler, Eric Vittinghof, 
and other users for sending feedback and bug reports that greatly
helped to improve \pname. 

The development of DAGitty was sponsored by funding from the 
Institute of Genetics, Health and Therapeutics at Leeds
University, UK. I thank George Ellison for arranging this
generous support.

\section{Legal notice}

Use of \pname is (and will always be) freely permitted and free of charge. 
You may download a copy of \pname's source code from its website at \url{www.dagitty.net}.
The source code is available under the GNU General Public License (GPL),
either version 2.0, or any later version, at the licensee's choice;
see the file \verb|LICENSE.txt| in the download archive for details.
In particular, the GPL permits you to  modify and redistribute the source
as you please as long as the result remains itself under the GPL.

\section{Bundled libraries}

\pname ships along with the following JavaScript libraries:

\begin{itemize} 
 \item \emph{Prototype.js}, a framework that makes life with JavaScript much easier. Only some 
  parts of Prototype (mainly those focusing on data structures) are included to keep the code small. Developed
  by the Prototype Core Team and licensed under the MIT license \cite{Prototype2010}. 
\end{itemize}

Furthermore, \pname uses some modified code from the \emph{Dracula Graph Library} by 
Philipp Strathausen, which is also licensed under the MIT license \cite{Dracula2010}. 

Versions of \pname prior to 2.0 used the \emph{Rapha\"el} library for smooth 
cross-browser vector graphics in SVG and VML, developed by Dmitry Baranovskiy 
\cite{Raphael2010}. However, the dependency on \emph{Rapha\"el} was removed starting
from version 2.0, and only SVG-capable browsers will be supported in the future.

I am grateful to the authors of these libraries for their valuable work. 

\section{Bundled examples}

\pname contains some builtin examples for didactic and illustrative purposes. 
Some of these examples are taken from published papers or talks given at 
scientific meetings. These are, in inverse chronological order: 

\begin{itemize}
 \item van Kampen 2014 \cite{Kampen2014} 
 \item Polzer et al., 2012 \cite{Polzer2012}
 \item Schipf et al., 2010 \cite{Schipf2010}
 \item Shrier \& Pratt, 2008 \cite{ShrierP2008}
 \item Sebastiani et al.\footnote{The example actually shows only a small part of their DAG.}, 
    2005 \cite{Sebastiani2005}
 \item Acid \& de Campos, 1996 \cite{Acid1996}
\end{itemize}

Another example was provided by Felix Thoemmes 
via personal communication (2013).

\section{Author contact}

I would be glad to receive feedback from those 
who use \pname for research or educational purposes. 
Also, you are welcome to send me your 
suggestions or requests for features that 
you miss in \pname. 

\bigskip

\noindent
Johannes Textor \\
Theoretical Biology \& Bioinformatics \\
Universiteit Utrecht, The Netherlands \\
\\
\href{mailto:johannes.textor@gmx.de}{\tt johannes.textor@gmx.de} \\
\href{http://theory.bio.uu.nl/textor/}{\tt theory.bio.uu.nl/textor/} \\
Twitter: \href{https://twitter.com/hashtag/dagitty}{\tt \#dagitty}

\bigskip

\bibliography{main}

\end{document}